\documentclass[10pt,twocolumn,letterpaper]{article}

\usepackage{cvpr}

\usepackage{ marvosym }
\def\MAIL{{\bf \Letter}}

\usepackage{epsfig}
\usepackage{graphicx}
\usepackage{amsmath}
\usepackage{amssymb}
\usepackage{bbm}
\usepackage[usenames, dvipsnames]{color}
\usepackage{multirow}
\usepackage{caption}
\usepackage{xspace}
\usepackage[pagebackref=true,breaklinks=true,letterpaper=true,colorlinks,citecolor=blue,linkcolor=blue,bookmarks=false]{hyperref}

\DeclareUrlCommand\url{\def\UrlLeft{$\sf   }\def\UrlRight{$}%
}

\cvprfinalcopy %

\def\OurMethod{{\rm MEInst}\xspace}
\def\bOurMethod{{\rm \bf MEInst}\xspace}

\ifcvprfinal\fi
\begin{document}

\title{Mask Encoding for Single Shot Instance Segmentation
}

\author{Rufeng Zhang$^1$, 
~
~
~
~
Zhi Tian$^2$, 
~
~
~
~
Chunhua Shen$^2$\thanks{Corresponding author, 
\MAIL \,
\href{mailto:chunhua.shen@adelaide.edu.au?Subject=Mask-Encoding-CVPR2020 Question}{\color{blue}{$\sf chunhua.shen@adelaide.edu.au$}}}
, 
~
~
~
~
Mingyu You$^1$,
~
~
~
~
Youliang Yan$^3$
\\[.152cm]
$ ^1 $ Tongji University, China ~ ~   
$ ^2 $ University of Adelaide, Australia
~ ~  
$ ^3 $ Huawei Noah's Ark Lab
}

\maketitle
\thispagestyle{empty}

\begin{abstract}
   To date,  instance segmentation is dominated by two-stage methods, as pioneered by Mask R-CNN. 
   In contrast, one-stage alternatives cannot compete with Mask R-CNN in mask AP, mainly due to the difficulty of compactly representing masks, making the design of one-stage methods very challenging.  
   In this work, we propose a 
   simple 
   single-shot instance segmentation framework, termed mask encoding based instance segmentation {\rm (\OurMethod)}. Instead of predicting the two-dimensional mask directly, \OurMethod distills it into a compact and fixed-dimensional representation vector, which allows the instance segmentation task to be incorporated into one-stage bounding-box detectors and results in a simple yet efficient instance segmentation framework.
   The proposed one-stage \OurMethod achieves $ 36.9 \%$ in mask AP with single-model (ResNeXt-101-FPN backbone) and single-scale testing on the MS-COCO benchmark. 
   We show that the much simpler and flexible one-stage instance segmentation method, can also achieve competitive performance.
   This framework can be easily adapted for other instance-level recognition tasks.
   
   Code is 
   available at:
   \href{https://git.io/AdelaiDet}{\color{blue}{$\tt git.io/AdelaiDet$}}
\end{abstract}

\section{Introduction}

Instance segmentation enables various visual applications like autonomous driving and robot navigation, to name a few. Instead of separately detecting objects or assigning category labels to pixels, instance segmentation unifies these tasks together, thus being one of the most challenging tasks in computer vision.

Recent advances in deep convolutional neural networks (CNNs) have enabled tremendous progress in instance segmentation, \eg,~\cite{he2017mask, huang2019mask, li2017fully, liu2018path}. One of the mainstream methods employs a two-stage pipeline that first generates proposals and then performs pixel classification within each proposal, as popularized by Mask R-CNN~\cite{he2017mask}. Almost all the methods in the top rank on the challenging COCO benchmark~\cite{lin2014microsoft} are built upon Mask R-CNN thus far. 
One drawback of these two-stage solutions is not sufficiently efficient as their runtime is constrained by the number of instances in an image. On the other hand, one-stage paradigms process the full image straightforward, making the speed stable no matter how many objects %
present.

\begin{figure}[t!]
    \centering
    \includegraphics[width=.9\linewidth]{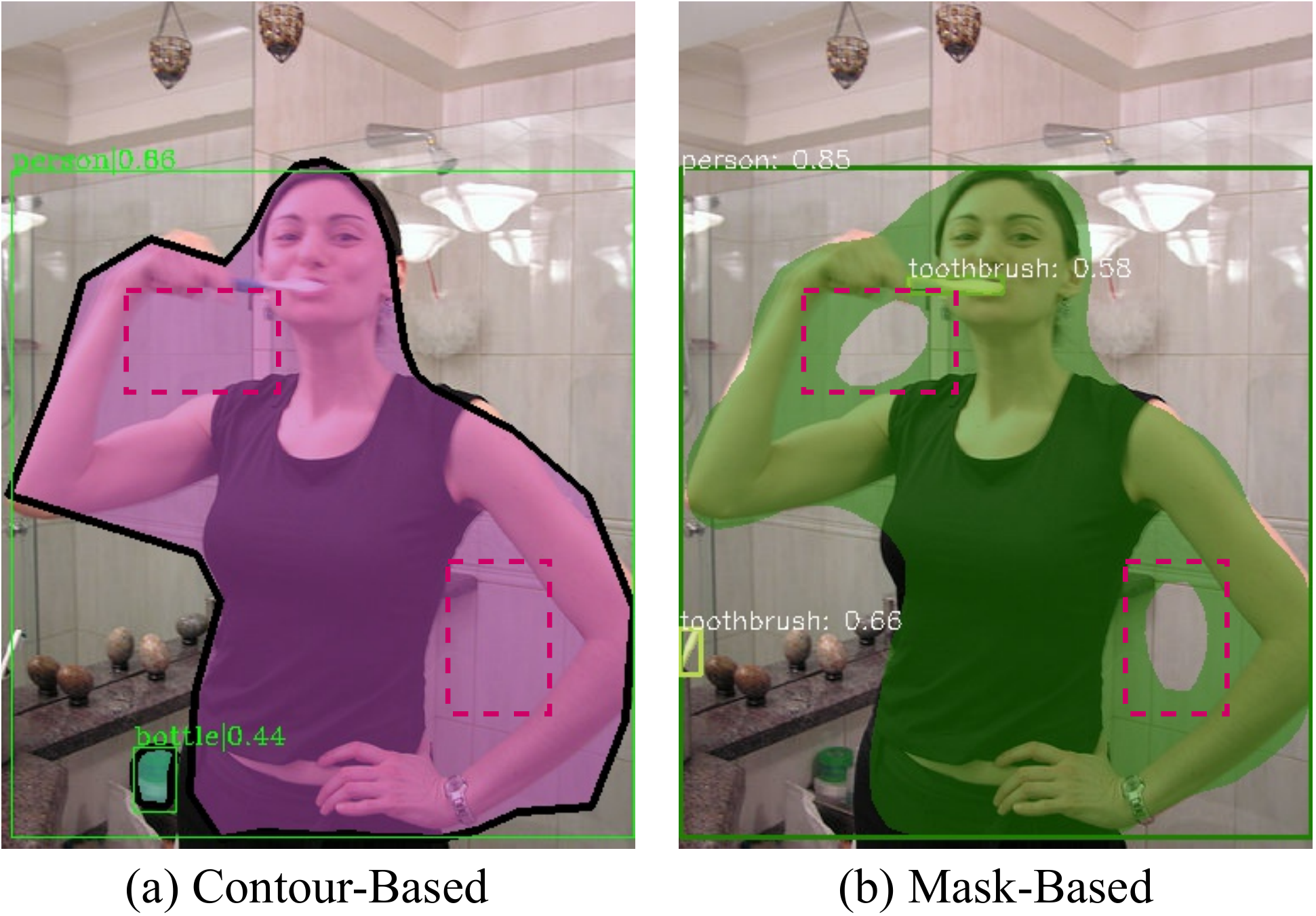}
    \caption{\textbf{Contour-Based~\cite{xu2019explicit} \vs Mask-Based.} 
    ``Hollow Decay" is depicted with \textbf{\color{red}red} dashed rectangles. The contour-based methods exhibit systematic artifacts on ``disjointed" objects. %
    }
    \label{fig:hollow}
\end{figure}

Several works have attempted to incorporate mask prediction into fully convolutional networks (FCNs)~\cite{long2015fully}, resulting in single shot instance segmentation frameworks. These algorithms share a common insight, \ie, encoding the object shape with a set of contour coefficients. 
Specifically, ESE-Seg~\cite{xu2019explicit} designs an ``inner-center radius" shape signature for each instance and fits it with Chebyshev polynomials.
Concurrently, PolarMask~\cite{xie2019polarmask} regresses the dense distance of rays between mass-center and contours. These contour-based methods enjoy the advantages of easy optimization and fast inference.
The major issue of these methods is that the predicted masks may exhibit 
``hollow decay" inevitably, since they can only depict instances with a single contour, as shown in Figure~\ref{fig:hollow}.

Alternatively, a non-parametric mask representation  
is more natural for mask prediction as traditionally done, with the price
of increasing both design and computation complexity. 
As natural object masks are not random and 
akin to 
natural images, instance masks reside in a much lower intrinsic dimension than that of the pixel space. 
This inspires us to ask a question, ``Is it possible to predict the object mask 
in the intrinsic low-dimensional space and still achieve %
competitive 
accuracy?"
Here we provide an affirmative answer:
we propose to  
encode instance masks 
using a
learned 
dictionary such that only a few scalar coefficients are needed to 
represent each mask. 
We demonstrate that such an approach  is robust to noise, and
 efficient, 
easy to decode 
for reconstruction.

Then an 
one-stage detector such as RetinaNet~\cite{lin2017focal}, FCOS~\cite{tian2019fcos} can be easily extended by adding a branch for predicting 
these fixed-dimensional mask 
coefficients,
along with the bounding box regression and category classification branches. 
We build our method on top of FCOS for its simplicity and good detection performance. 

We demonstrate that 
our method %
can 
outperform recent one-stage algorithms 
\cite{bolya2019yolact, xie2019polarmask, xu2019explicit, zhou2019bottom} with 
this 
simple design.
In particular, experiments on the COCO \texttt{val2017} show that \OurMethod 
achieves 
a large gain compared to ESE-Seg~\cite{xu2019explicit}, 
outperforming by 
$11.8\%$ in AP$_{50}$ and $16.5\%$ in AP$_{75}$, respectively. 
Our model 
beats 
PolarMask~\cite{xie2019polarmask} in accuracy with similar computational complexity, owing to the lower reconstruction error and 
more effective 
reconstruction. %
This is expected, as the mask representation of our method is more powerful 
than the parametric representation of~\cite{xie2019polarmask, xu2019explicit}.

Additionally, we 
take a closer look at how the object detector influences the performance of instance segmentation based on extensive qualitative experiments. With a careful design based on our finding, \OurMethod achieves comparable performance with Mask R-CNN~\cite{he2017mask} with the advantage of being much simpler and flexible.

It is noteworthy that  our method is compatible with 
most 
one-stage detection frameworks including the anchor-free paradigm. We demonstrate its generality using the FCOS detector, and evaluate the performance on the %
COCO benchmark~\cite{lin2014microsoft}. Other anchor-based methods such as YOLO~\cite{redmon2016you}, RetinaNet~\cite{lin2017focal} 
may be used here 
with
minimum
modification. Moreover, the vanilla detectors can also benefit from the paralleled mask prediction branch, 
improving the bounding box detection 
accuracy.

The main contributions of this work can be summarized as follows.

\begin{itemize}
\itemsep -2pt
    \item We propose to encode a  two-dimensional instance mask into a compact representation vector. 
    The compressed vector, takes advantages of the redundancy in the original mask and proves to be effective and efficient for reconstruction.
    
    Encoding can be done with a few  dictionary learning methods, including PCA, sparse coding, and auto-encoders. Here we show that even the simplest 
    PCA already suffices for mask encoding. 
    
    \item 
    With this mask representation, 
    a new  framework is introduced for single shot instance segmentation, termed
    mask encoding based instance segmentation 
    (\OurMethod), by extending FCOS~\cite{tian2019fcos} with a mask branch for 
    mask coefficient 
    regression. Actually, our mask encoding is completely independent of the mechanism of detectors, and it may be easily incorporated into 
    other 
    detectors. 

    \item 
    We demonstrate a simple and flexible one-stage instance segmentation method.
    Our best model, attains mask AP of $38.2\%$ on COCO \texttt{test-dev}, achieving 
    a good balance between 
    accuracy and speed. 
\end{itemize}

\section{Related Work}
We review a few works that are most relevant to ours.

\textbf{Two-stage Instance Segmentation}
The mainstream approaches to instance segmentation~\cite{dai2016instance, he2017mask, huang2019mask, liu2018path} inherit the pipeline 
of 
two-stage object detectors, as pioneered by Mask R-CNN~\cite{he2017mask}. 
These methods typically detect instance bounding boxes and then perform binary-class segmentation in boxes.
Compared with segmentation-driven ones~\cite{arnab2017pixelwise, liu2017sgn}, this 
group of
paradigms lead on
most benchmarks~\cite{cordts2016cityscapes, lin2014microsoft} in accuracy.
In particular, Mask R-CNN~\cite{he2017mask} replaces ROIPool with ROIAlign to 
better align features.
Following Mask R-CNN, Liu \etal~\cite{liu2018path} present bottom-up path augmentation and adaptive feature pooling for further feature optimization. 
Mask Scoring R-CNN~\cite{huang2019mask} extends Mask R-CNN with an extra MaskIoU branch, aiming to calibrate the mismatch between mask's quality and the corresponding confidence. The above methods consistently advance the performance. 

\textbf{One-stage Instance Segmentation} 
The second family of solutions~\cite{arnab2017pixelwise, bai2017deep, kirillov2017instancecut, liu2017sgn} are 
built upon the success of semantic segmentation, \ie, generating pixel-wise classification maps firstly and then clustering them into instances. Specifically, InstanceCut~\cite{kirillov2017instancecut} 
addresses  the problem with two paralleled sub-tasks, instance-agnostic segmentation and instance-specific boundaries. 

In the meantime, dense object segmentation has not witnessed remarkable progress. %
Impressively, several works have attempted to fill in the gaps lately. 
For example,
TensorMask~\cite{chen2019tensormask} can be viewed as a precursor to this group of algorithms, in which a structured 4D tensor is introduced to represent the mask over a spatial domain. It achieves similar performance with two-stage methods 
with the cost of heavy computation overhead in training and testing. 
In YOLACT~\cite{bolya2019yolact}, a series of global prototypes and individual linear coefficients are assembled for masks, achieving a real-time speed.
BlendMask \cite{blendmask} improves YOLACT in both accuracy and speed.
Recently, Xie \etal propose a general framework named PolarMask~\cite{xie2019polarmask}, which is capable to directly predict the mask without bounding box using a parametric representation of masks. 
More recently, SOLO and its improved version SOLOv2 demonstrate promosing results with a simple
FCN-like framework \cite{SOLO, SOLOv2}.

\begin{figure*}[!htp]
    \centering
    \includegraphics[width=1\linewidth]{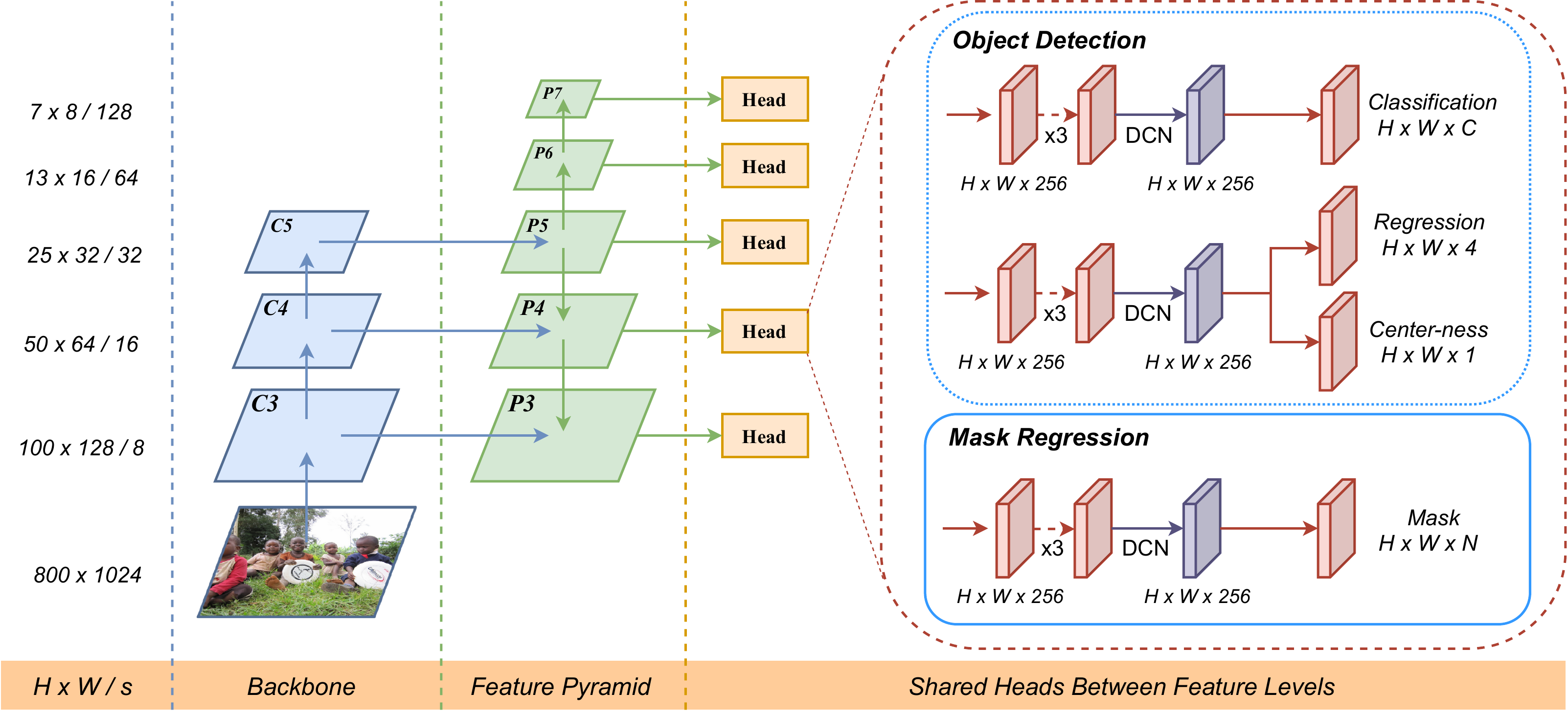}
    \caption{\textbf{The overall architecture of \bOurMethod}, which extends FCOS~\cite{tian2019fcos} with a Mask  Regression Branch. The model mainly consists of four modules: (a) Backbone %
    for feature extraction. (b) Feature Pyramid. 
    (c) Detection Heads for object detection. (d) Mask Regression Branch for instance segmentation.
    \OurMethod detects objects and predicts their mask vectors simultaneously, in which the first three processes are consistent with FCOS. Then the instance masks are reconstructed efficiently through Eq.~\eqref{equ_mapping} (right). Here  DCN denotes deformable convolution,
    which is optional (best viewed in color), and \textit{N} means the dimension of representation vectors (\eg, $N=60$). %
    }
    \label{fig:Network_architecture}
\end{figure*}

\section{Our Method}

In this section, we first %
present 
the overall architecture of \OurMethod.
We then introduce the instance representation with mask encoding and its optimization. Finally, we explore the correlation between detection quality and mask generation to further improve the performance of \OurMethod.

\subsection{Network Architecture}

The object detection modules in our method mainly inherit the pipeline from FCOS\footnote {We use the improved version, including sharing the features between center-ness and regression branch, central sampling and so on. Please refer to ~\cite{tian2019fcos} for further details.}~\cite{tian2019fcos} for its flexibility and simplicity, including a backbone module~\cite{he2016deep}, a feature pyramid module~\cite{lin2017feature}, and two task-specific heads for classification, box regression and center-ness (they share the same head). 
Then a parallel branch is 
included 
for predicting encoded mask coefficients. 
Additionally, 
we carefully re-design some parts of the framework,
which further boosts the performance. Details are  discussed in the following subsection. The overall framework is illustrated in Figure~\ref{fig:Network_architecture}.

\subsection{Mask Encoding}

\begin{figure}[t]
\begin{center}
\includegraphics[width=0.9\linewidth]{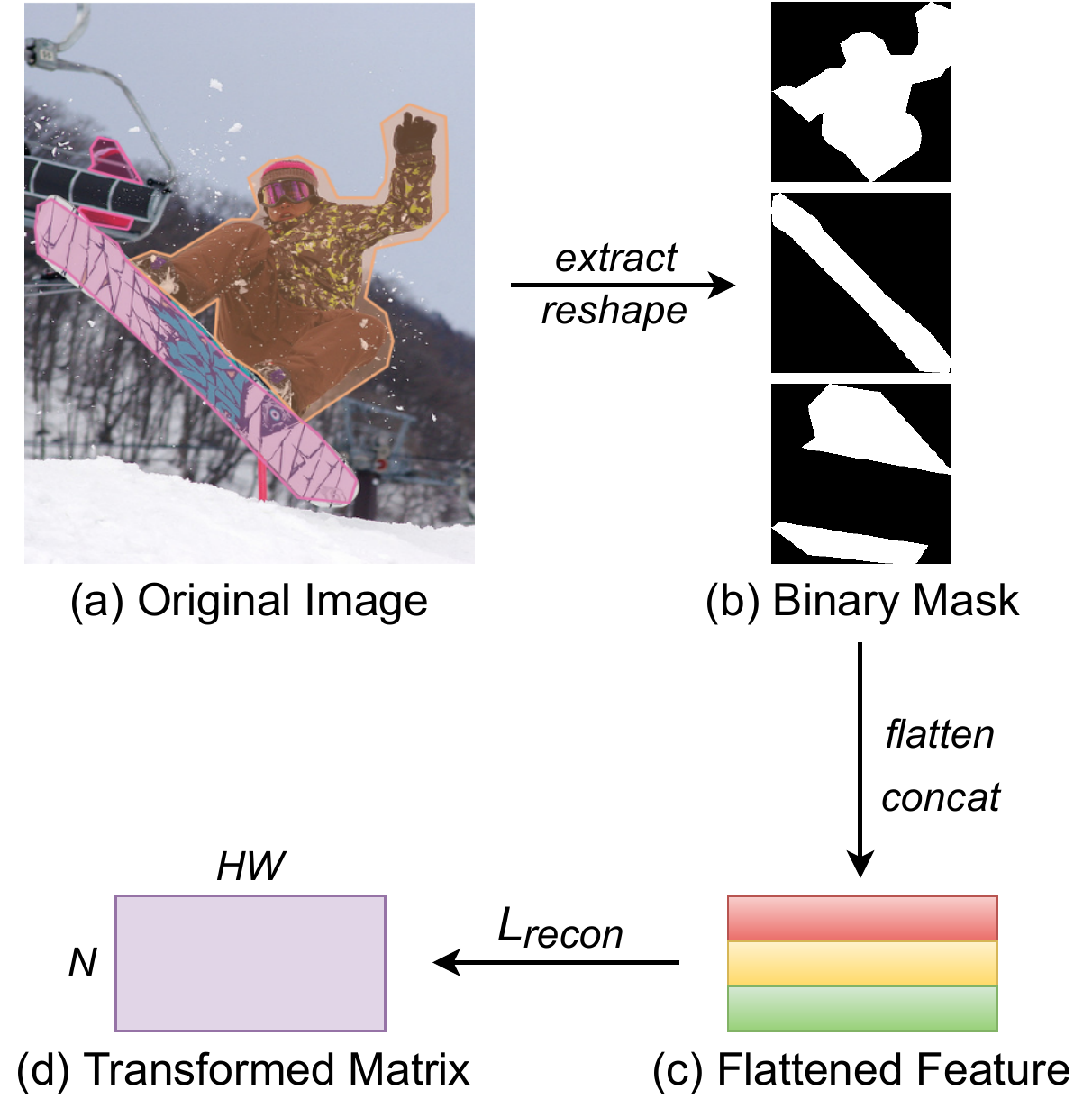}
\end{center}
   \caption{\textbf{The pipeline of mask encoding}. 
   (a) is the original image annotated with instance labels. We extract these annotations and reshape them as (b) $m \times m$ mask (here mask is class-agnostic). Then (c) the flattened feature is compressed for dimensionality reduction. Finally we harvest (d) transformed matrix for mask encoding. 
   The entire procedure is done off-line and it performs very fast. 
   After learning,
   we freeze all these parameters during network training and inference.
   }
\label{fig:pca_process}
\end{figure}

Given a structured instance mask, we can easily figure out the redundancy in its representation.
An example can be seen in~Figure~\ref{fig:pca_process}(b). The discriminative pixels are mainly distributed along the object boundaries while most pixels in its body hold the properties of being  category-continuous and category-consistent. In other words, the existing mask representations contains redundant information and %
it may be highly compressed with 
negligible 
loss. In this subsection, we 
describe how to encode the two-dimensional geometry into a much more compact representation vector in detail.

\def\p{{\boldsymbol{ v  }}}
\def\m{{\boldsymbol{ u }}}

\textbf{Compact Representation}
Let $\mathbf{M}'\in\mathbb{R}^{H \times W}$ represent the ground truth mask and ${\p}\in\mathbb{R}^{N}$ be the compressed vector, where $H$, $W$ and $N$ denotes the height/width of two-dimensional mask and the dimension of compact representation vector, respectively. Typically $ N \ll  H \cdot W $. 
Note that here $\mathbf{M}' $ is class-agnostic and therefore all the categories are encoded with binary-class encoding, \ie, $\mathbf{M}' \in\{0,1\}^{H \times W}$. The mask is flattened to be a vector for ease of calculation, as ${\m}\in\mathbb{R}^{HW }$. 
In order to compress $\m $ into $\p$, we seek a transformation under some criterion to minimize the reconstruction error between $\m$ and $\p $.
Although many approaches can be used for our purpose here,
we observe that the simple linear projecting can already perform well in our experiment. 
In particular, we have,
\begin{equation} \label{equ_mapping}
\begin{aligned}
\p  &= \mathbf{T} \m ; & {\tilde{ \m }} &= \mathbf{W} \p.
\end{aligned}
\end{equation}
Here 
$\mathbf{T}\in\mathbb{R}^{N \times HW}$ is the project matrix, used to 
compress 
$\m $ into $\p $. 
$\m $ can be recovered with the reconstruction matrix $\mathbf{W}\in\mathbb{R}^{HW \times N}$.
Note that, 
$ \m $ is centered by subtracting its mean over the training set, followed with normalization. 
Finally, we obtain these matrices by minimizing the reconstruction error between $\m $ and $\tilde{\m}$  on the training set. 
Mathematically it is written as Eq.~\eqref{equ_err_reconstruction}:
\begin{equation} \label{equ_err_reconstruction}
\begin{aligned}
\mathbf{T^{\ast}, W^{\ast}} &= \mathop{\arg\min}_{\mathbf{T,W}}\sum_{\m}\|\mathbf{ \m -\tilde{\m}}\|^{2} \\
&= \mathop{\arg\min}_{\mathbf{T,W}}\sum_{ \m }\| {\m - {\bf W T} \m }\|^{2}
\end{aligned}
\end{equation}

We follow the strategy in DUpsampling~\cite{tian2019decoders} and 
optimize 
this objective by using principal component analysis (PCA).
The overall process is illustrated in Figure~\ref{fig:pca_process}. 
Please refer to  DUpsampling~\cite{tian2019decoders} for details.
There may be %
alternative 
options to minimize the reconstruction loss, \eg, sparse coding or non-linear auto-encoder.
\textbf{Mask Reconstruction} Given the predicted representation vector ${\hat{ \p }}\in\mathbb{R}^{N}$, the two-dimensional mask $\mathbf{{M'}}\in\mathbb{R}^{H \times W}$ can be 
reconstructed through Eq.~\eqref{equ_mapping} (right). As we employ this operation after non-maximum suppression (the highest scoring 100 samples), the computation cost of such matrix multiplication is 
negligible.

\textbf{Loss Function}
We define our mask loss function as follows:
\begin{equation}
    \label{equ_loss}
    \mathcal{L}_{\mathit{mask}} = \mathbbm{1}^{\mathit{obj}}\sum_{i}^{N}
    d_{mask}(\hat{y}_{i}, y_{i}),
\end{equation}
where 
$\mathbbm{1}^{\mathit{obj}}$ is the indicator function for positive samples. $\hat{y}_{i}$, $y_{i}$ denotes the $i$-th element in prediction and ground-truth vectors, respectively. 
In our implementation, we have compared different forms of $d_{mask}(\cdot, \cdot)$, \eg, $l_1$ loss, smooth-$l_1$ loss, $l_2$ loss and cosine similarity loss. Finally, we employ $l_2$ loss for its effectiveness and stability in training.
We append it to the 
overall 
loss, formally,
\begin{equation}
    \label{total_loss}
    \mathcal{L} = \lambda_{det}  \cdot  \mathcal{L}_{\mathit{det}} + \lambda_{mask}
    \cdot 
    \mathcal{L}_{\mathit{mask}}.
\end{equation}
Here 
$\mathcal{L}_{\mathit{det}}$ is the loss for detection, consisting of $\mathcal{L}_{\mathit{cls}}$ for classification, $\mathcal{L}_{\mathit{reg}}$ for bounding box regression and $\mathcal{L}_{\mathit{cen}}$ for center-ness. In particular, $\mathcal{L}_{\mathit{cls}}$ is focal loss as in~\cite{lin2017focal}, $\mathcal{L}_{\mathit{reg}}$ is the GIoU loss following 
FCOS \cite{tian2019fcos}.
$\mathcal{L}_{\mathit{cen}}$ denotes the binary cross entropy (BCE) loss
for center-ness. All the balance weights in $\mathcal{L}_{\mathit{det}}$ are set to $1$ for simplicity in our experiments.

\begin{table*}[t]
\small 
\centering 
\begin{tabular}{ c |c|c c c|c c c}
 Detector & AP & AP$^{bb}$ & AP$^{bb}_{50}$ & AP$^{bb}_{75}$
 & AP$^{bb}_{S}$ & AP$^{bb}_{M}$ & AP$^{bb}_{L}$   \\
 \hline
 Mask-R-50-FPN & 34.2 & 37.8 & 59.3 & 41.1 & 21.5 & 41.1 & 49.9 \\
 FCOS-R-50-FPN & 34.1{\color{red}($-$0.1)} & 38.7{\color{OliveGreen}(+0.9)} & 57.3{\color{red}($-2$)} & 41.9{\color{OliveGreen}(+0.8)} & 22.6{\color{OliveGreen}(+1.1)} & 42.4{\color{OliveGreen}(+1.3)} & 50.1{\color{OliveGreen}(+0.2)} \\
 \hline
 Mask-R-101-FPN & 35.7 & 40.1 &	61.7 & 44.0 & 23.1 & 43.4 & 52.7 \\
 FCOS-R-101-FPN & 36.6{\color{OliveGreen}(+0.9)} & 42.9{\color{OliveGreen}(+2.8)} & 61.8{\color{OliveGreen}(+0.1)} & 46.3{\color{OliveGreen}(+2.3)} & 27.4{\color{OliveGreen}(+4.3)} & 46.9{\color{OliveGreen}(+3.5)} & 55.4{\color{OliveGreen}(+2.7)} 
 \\
 \hline
 Mask-X-101-32x8d-FPN & 36.9 & 42.2 & 63.9 & 46.1 & 25.4 & 46.1 & 54.7 \\
 FCOS-X-101-32x8d-FPN & 37.1{\color{OliveGreen}(+0.2)} & 44.0{\color{OliveGreen}(+1.8)} & 63.2{\color{red}($-$0.7)} & 47.6{\color{OliveGreen}(+1.5)} & 27.5{\color{OliveGreen}(+2.1)} & 47.6{\color{OliveGreen}(+1.5)} & 56.4{\color{OliveGreen}(+1.7)} 
\end{tabular}
\caption{Comparisons among different algorithms on the COCO \texttt{val2017} split. The first row shows Mask R-CNN~\cite{he2017mask} trained by He \etal, while the other is FCOS~\cite{tian2019fcos} with the same backbone network. We 
only
employ them to detect objects, as for Mask R-CNN, we 
discard
the mask outputs. AP indicates the performance of instance segmentation, which is predicted by the same model with different pre-detected boxes. The gap between two detectors are highlighted by \textbf{\color{OliveGreen}green} and \textbf{\color{red}red}, respectively. \textbf{\color{OliveGreen}green} means \textbf{better} and \textbf{\color{red}red} \textbf{worse}.
}
\label{table:Correlation}
\end{table*}

\begin{figure}[t!]
\centering
\includegraphics[trim =0mm 0mm 0mm 0mm, clip, width=1\linewidth]{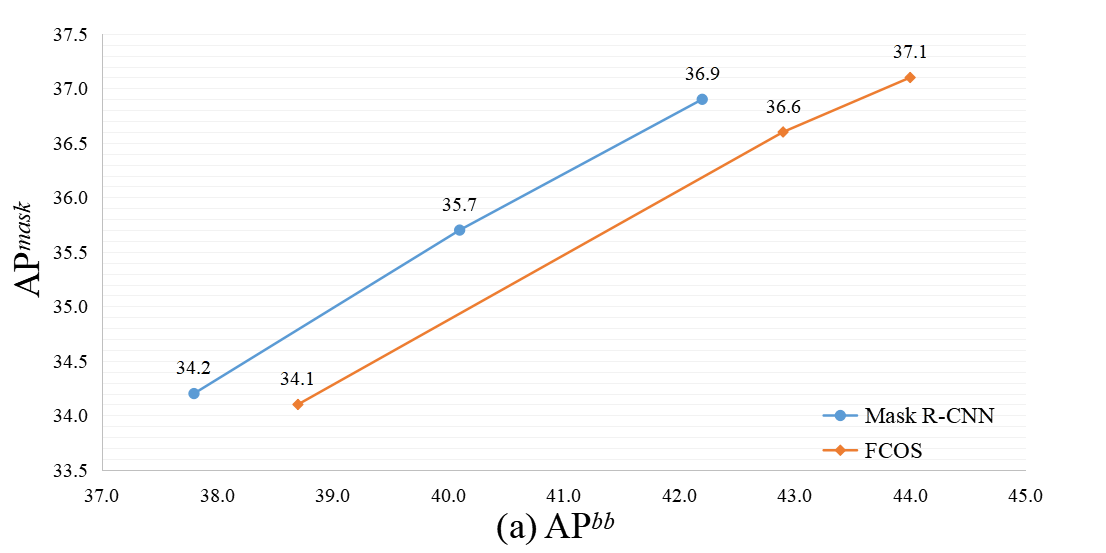}
\includegraphics[trim =0mm 0mm 0mm 0mm, clip, width=1\linewidth]{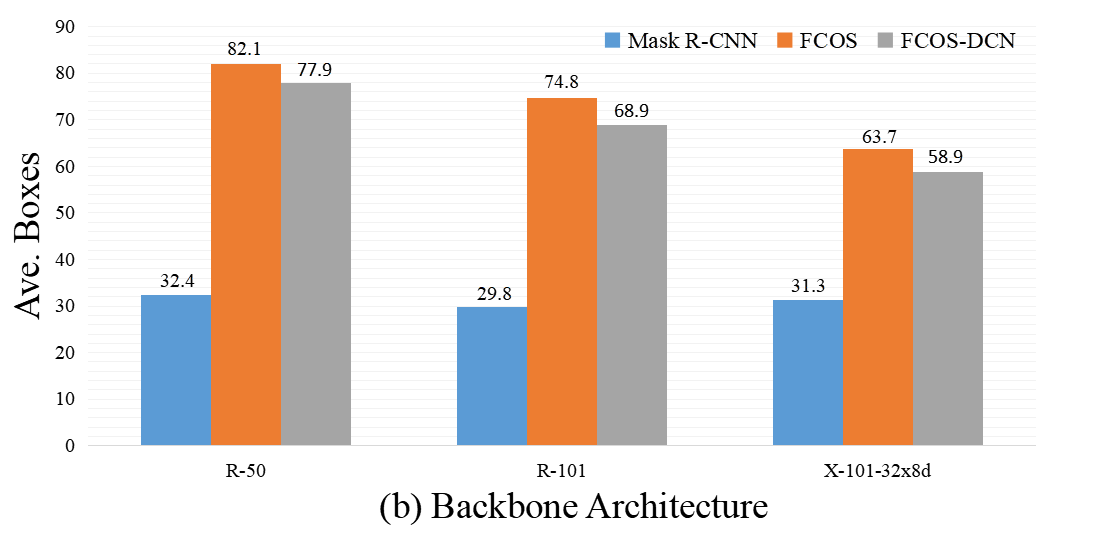}
\caption{\textbf{Quantitative analysis of different paradigms} on the COCO \texttt{val2017} split. 
(a) {AP$^{bb}$ \vs\ AP$^{mask}$,} which shows the correlation between box and mask. 
As for the same pipeline, better detectors lead to 
better performances in instance mask. 
However, 
this is not the case for FCOS, whose overall detection result is better than the corresponding Mask R-CNN. But FCOS only  performs similar or even worse in instance segmentation. 
(b) {Backbone architecture} \vs\ 
{Average number of boxes per image:}
Compared with Mask R-CNN, FCOS outputs more than $2$ times more boxes, resulting in lower  AP$^{bb}_{50}$.
The phenomenon
can be alleviated 
with a larger receptive field.   
}
\label{fig:Correlation}
\end{figure}

\begin{figure}[t!]
\centering
\includegraphics[width=1\linewidth]{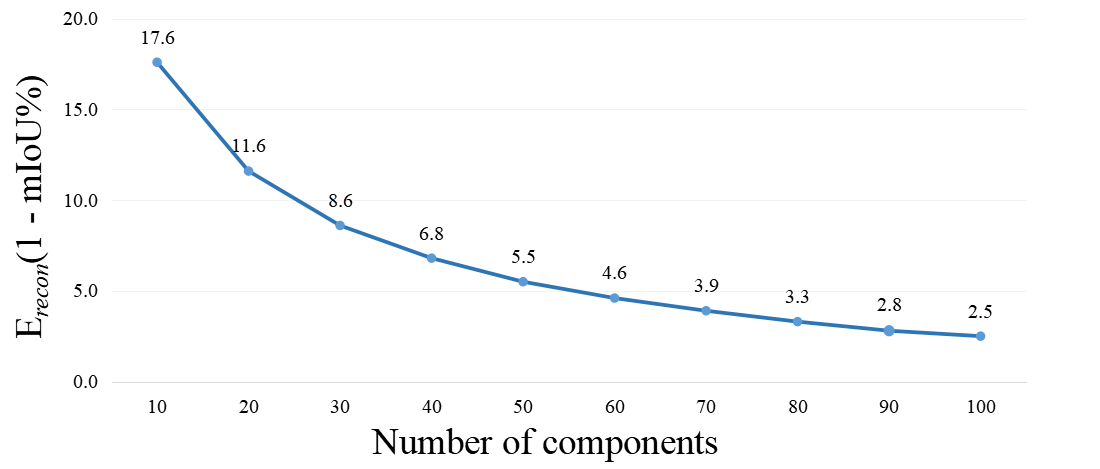}
\caption{\textbf{The reconstruction error} $E_{recon}$ \vs\ Number of components to keep on COCO \texttt{train2017} split.}
\label{fig:reconstruction_err}
\end{figure}

\begin{figure*}[!htp]
    \centering
    \includegraphics[width=1\linewidth]{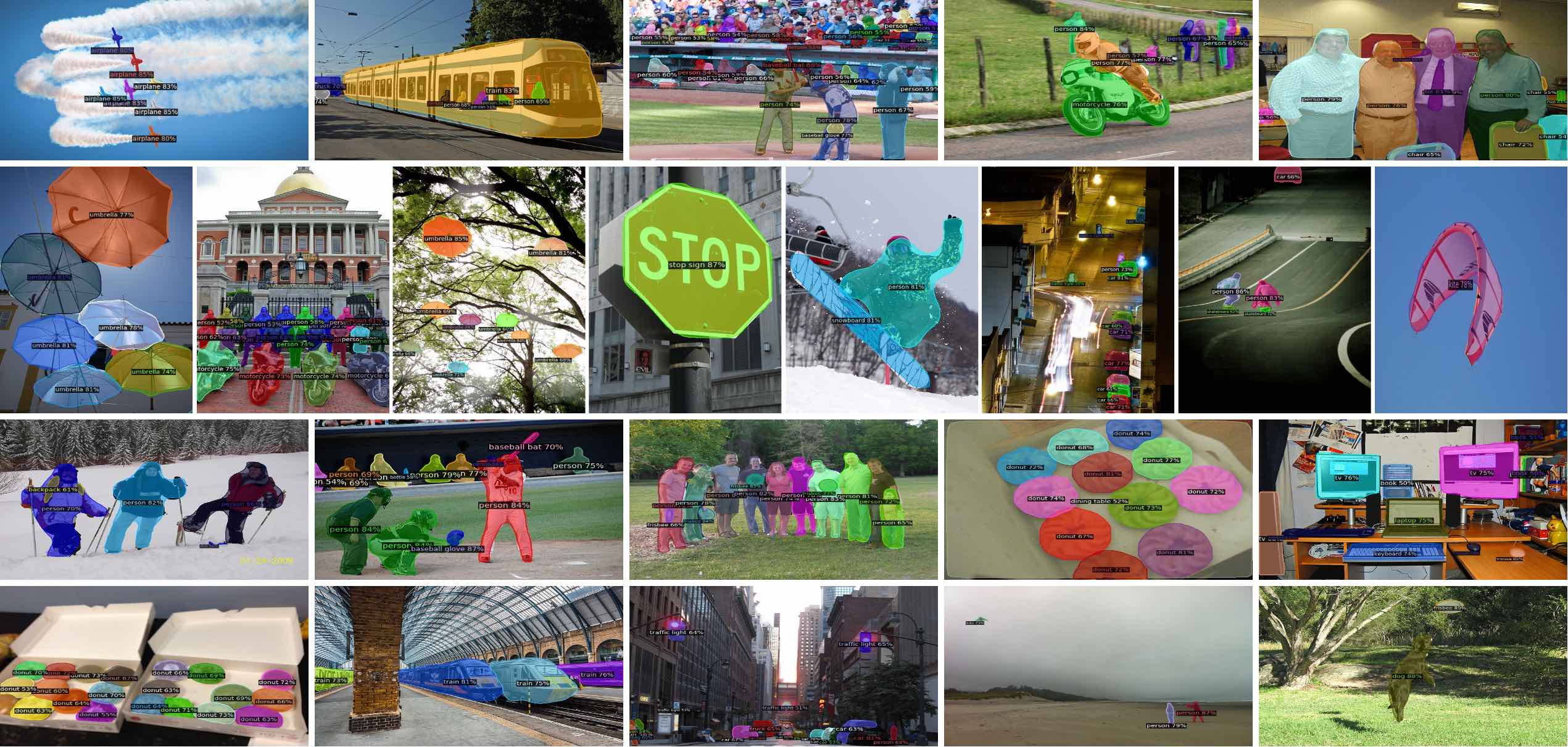}
    \caption{\textbf{Visualization of \OurMethod on COCO images} with ResNeXt-101-FPN, achieving $36.9\%$ mask AP (Table~\ref{table:SOTA}).}
    \label{fig:Visualization}
\end{figure*}

\subsection{Correlation Between Boxes and Masks}

In general, instance segmentation and object detection are inseparable in detection-driven pipelines. Intuitively, better bounding boxes 
improves 
the overall performance in the mask branch. Here we carry out several experiments to validate our assumptions empirically.

Take Mask R-CNN~\cite{he2017mask} as an example. The inference flow is as follows:
1) A backbone module is used to extract semantic feature from the input image.
2) The extracted feature is then sent to the following modules for classification and object regression.
3) Afterwards, the mask stage computes features using ROIAlign from each detected box.
4) Finally, the regional representation is performed pixel-wise segmentation. It only predicts a binary mask.

In our experiments, the Mask-R-50-FPN model pre-trained by He \etal is %
used 
as the main backbone. 
The \textit{step-2} in the above process is replaced with a series of pre-acquired detection results predicted by different detectors, in which case all the variables are kept 
the same 
except the boxes. 
Here we choose Mask R-CNN~\cite{he2017mask} (two-stage) and FCOS~\cite{tian2019fcos} (one-stage) with different backbones as object detectors. In the sequel, AP means  \textit{mask} AP and \textit{box} AP is denoted as AP$^{bb}$. The quantitative results are  shown in Table~\ref{table:Correlation} and Figure~\ref{fig:Correlation}.

As for the same architecture, the detector brings consistent and noticeable gain in mask when the network goes deeper. However, the results of instance segmentation fall below our expectations with different pipelines. Compared with Mask R-CNN, FCOS achieves better detection performances among all backbones under the metric AP$^{bb}$, measuring $0.9\%$, $2.8\%$, $1.8\%$, respectively. Nevertheless, the corresponding segmentation has not been witnessed equivalent 
improvement, 
and even performs worse ($34.1\%$ \vs\  $34.2\%$).
It seems counter-intuitive.

We observe that FCOS performs better under all the general metrics except {AP$^{bb}_{50}$}, which indicates that the boxes predicted by FCOS are location-accurate but with more false-positive (FP).
Figure~\ref{fig:Correlation}(b) shows the average number of bounding boxes predicted by different models. 
FCOS predicts significantly more bounding boxes than Mask R-CNN with the same confidence threshold (\eg, $0.05$), which may degrade the performance under the metric AP$^{bb}_{50}$. 
Mask R-CNN employs a two-stage pipeline, \ie, first proposes candidates and then refines the boxes, in which case most mis-proposed boxes can be filtered out effectively. However, one-stage paradigm such as FCOS outputs results directly for faster inference, resulting in the redundant boxes. Actually almost all the one-stage methods~\cite{lin2017focal, liu2016ssd, redmon2017yolo9000} suffer from this dilemma. 

We hypothesize that the issue may be related to the effective receptive field (ERF). 
Zhou \etal~\cite{zhou2014object} declare that the effective receptive field is much smaller than the theoretical receptive field, since CNN tends to capture information from central regions. 
The insufficient ERF may lead to many false-positive (FP) boxes as the network can not ``see" the  objects. 
To tackle this issue, we simply employ deformable convolution~\cite{zhu2019deformable} that has the capacity to focus on salient regions and enlarge the ERF to some extent. 
Specifically, we replace the last vanilla convolutional layer in multi-head branches respectively. Note that other modules such as dilated convolution~\cite{chen2017deeplab} and Large Kernel~\cite{peng2017large}, which are beneficial to ERF, 
may 
also boost the performance. 
We 
provide 
further 
comparisons in the experimental section.
\section{Experiments}

Our experiments are conducted on the challenging MS COCO benchmark~\cite{lin2014microsoft} using the standard metrics for instance segmentation. All models are trained on the COCO \texttt{train2017} split ($\sim$118k images) and evaluated with \texttt{val2017} (5k images). The final results are reported on \texttt{test-dev} (20k images). Moreover, we adopt the $1\times$ training strategy~\cite{chen2019mmdetection, girshick2018detectron}, single scale training and testing unless otherwise specified. 

\textbf{Training Details} ResNet-50~\cite{he2016deep} is used as the backbone network and all hyper-parameters are kept consistent with FCOS~\cite{tian2019fcos} unless specified. Specifically, we use the stochastic gradient descent (SGD) optimizer, weight decay 0.0001, momentum 0.9 with 90K iterations in all. The initial learning rate is set to $0.01$  and divided by 10 at iteration 60K and 80K, respectively. We use a mini-batch of 16 images and all models are trained with 8 GPUs.
The backbone is initialized with the pre-trained weights on ImageNet~\cite{deng2009imagenet} and other newly added layers are initialized as in ~\cite{lin2017focal}. The shorter side of images is fixed as 800 pixels with the longer side being 1333 or less. Moreover, we sum up all the losses directly, \ie, $\lambda_{det} = \lambda_{mask} = 1$ in Eq.~\eqref{total_loss}. We expect 
that the performance may be better with a careful parameter tuning.

\textbf{Inference Details} The  inference process is kept the same as FCOS since we only append one more prediction to the predicted boxes. 
An input image goes through the network and then predicts boxes with several attributes, such as categories and mask coefficients.
We peform mask reconstruction after non-maximum suppression (NMS) to avoid unnecessary computational overhead (the highest scoring 100 samples). Since the matrix multiplication is %
fast, \OurMethod %
introduces slight 
overhead to its FCOS counterpart.

\subsection{Ablation Study}

\textbf{Analysis of Upper Bound} 
We first reshape all the annotations into $28 \times 28$ binary-class masks. Afterwards, these masks are encoded and recovered to two-dimensional matrices
with Eq.~\eqref{equ_mapping}. Finally we use the metric of \textit{mIoU} to evaluate the quality of reconstructed masks. The reconstruction error on 
the
COCO \texttt{train2017} split is shown in Figure~\ref{fig:reconstruction_err}. It is evident that the reconstruction error 
goes
down consistently with the increase of the number of components kept, and can even reach an extremely low level when the dimension goes to 100 (only $2.5\%$).
Moreover, we observe that the class-agnostic matrix achieves a similar result to class-specific one (up to $C$ times in dimensions). 
Thus, the former is a better choice for memory-conserving consideration.

\textbf{Dimension of Encoding Representation} 
It plays a very fundamental role in 
\OurMethod. As shown in Table~\ref{table:results_dimension}, the performance grows steadily with the increase of dimension and reaches saturation at last. For example, 
there is an improvement of $2\%$ from 20 to 60 and it remains 
stable 
beyond 
60. The reconstruction has a great influence at the beginning.
However, when adequate components can reconstruct the mask well,
it is no longer the main factor constraining the performance.
We choose $N = 60$ in our experiments unless otherwise specified.

\textbf{Learning without Explicit Encoding} 
Alternatively, the mask can be learned without explicit encoding. That is,
instead of compressing the redundant label into a fix-dimensional vector, we recover the predicted mask with the reconstruction matrix $\mathbf{W}$ and perform pixel-wise classification on it. This projecting process is essentially identical to employing a $1 \times 1$ convolution along the spatial dimensions, with convolutional kernels stored in $\mathbf{W}$. Note that these parameters are frozen during training. Moreover, we also explore the potential of learning without mask encoding, \ie, the network straightly outputs the high-dimensional masks (\eg,  $28 \times 28 = 784$). The results are shown in Table~\ref{table:results_PCA}. The over-high dimension makes it hard to optimize, resulting in a performance drop. Particularly, AP$_{75}$ and AP$_{L}$ decrease considerably, measuring by $1.3\%$ and $2.0\%$, respectively. The relatively compact vector
is not only for faster inference, but also beneficial for optimization. 
With the same dimension, our method still performs better under all the metrics, which further proves the effectiveness of mask encoding.   

\textbf{Loss Function}
As discussed above, mask encoding converts the task of instance segmentation into a set of coefficient regression problems. We try several popular losses in our experiments to
supervise the regression problems, more specifically, smooth-$l_1$ loss, $l_1$ loss and $l_2$ loss. $\lambda_{mask}$ in Eq.~\eqref{total_loss} is set to 1 for simplicity.
As shown is Table~\ref{table:loss}, $l_2$ loss performs better than others.
We also consider the case to view the mask vector as a whole, so we apply cosine similarity loss. However, the performance goes worse, which indicates that mask encoding has already eased the redundancy in original representation, and now the elements in vectors are independent.    

\textbf{Large Receptive Field}
Here we demonstrate the importance of large receptive field. 
Firstly, we apply large kernel~\cite{peng2017large} (LK) in the mask prediction layer. 
The LK layer is a combination of $1 \times k + k \times 1$ and $k \times 1 + 1 \times k$ convolutions. $k$ is set to 9 in our experiments. Compared with $3 \times 3$ convolution, it introduces 
negligible 
overhead. As shown in Table~\ref{table:large}, LK in prediction layer 
achieves
0.7\% AP gains. 
We also explore the potential of deformable convolution (DCN). Specifically, we %
only
use it in the last layer of head to keep our model efficient.
With the ability of capturing more meaningful and larger receptive features, it obtains 1.5\% improvement in AP. 

\begin{table}[t!]
\small
\centering
\begin{tabular}{c|c c c|c c c} 
$ N $  & AP & AP$_{50}$ & AP$_{75}$ & AP$_{S}$ & AP$_{M}$ & AP$_{L}$ \\
 \hline
 20 & 29.8 & 52.4 &	30.2 & 14.5 & 32.0 & 43.0 \\
 40 & 31.4 & 53.3 &	32.5 & 14.6 & 34.0 & 44.9 \\
 60 & 31.8 & \textbf{53.9} & \textbf{32.9} & \textbf{15.9} & 34.2 & \textbf{45.7} \\
 80 & \textbf{31.9} & \textbf{53.9} &	32.6 & 15.4 & \textbf{34.4} & 45.5 \\
\end{tabular}
\caption{\textbf{Number of components:} \OurMethod
attains
consistent gain with more components and reaches saturation at last.}
\label{table:results_dimension}
\end{table}

\begin{table}[t]
\small
\centering
\begin{tabular}{c|c c c|c c c} 
 encoding & AP & AP$_{50}$ & AP$_{75}$ & AP$_{S}$ & AP$_{M}$ & AP$_{L}$ \\
 \hline
 \checkmark & \textbf{31.8} & \textbf{53.9} & \textbf{32.9} & \textbf{15.9} & \textbf{34.2} & \textbf{45.7} \\
 $-$   & 30.8 &	53.3 & 31.6 & 14.5 & 33.1 &	43.7 \\
 w/o   & 29.7 & 52.7 & 29.9 &	14.5 & 32.0 & 43.4
\end{tabular}
\caption{\textbf{Mask encoding:} Learning with mask encoding achieves a better performance. 
Note that, 
the 
difference between ``$-$" and ``w/o" is that, the former one leverages implicit mask encoding, while the other does not.
}
\label{table:results_PCA}
\end{table}

\begin{table}[t]
\small
\centering
\begin{tabular}{c|c c c|c c c} 
 loss & AP & AP$_{50}$ & AP$_{75}$ & AP$_{S}$ & AP$_{M}$ & AP$_{L}$ \\
 \hline
 smooth $l_1$ & 30.8 & 53.2 & 31.5 & 14.8 &	33.0 & 44.7 \\
 $l_1$ & 31.4 &	53.4 & 32.4 & 15.3 & 33.8 &	44.8 \\
 $l_2$ & \textbf{31.8} & \textbf{53.9} & \textbf{32.9} & \textbf{15.9} & \textbf{34.2} & \textbf{45.7} \\
 cosine & 28.9 & 51.1 &	29.1 & 13.1 & 30.5 & 42.8  
\end{tabular}
\caption{\textbf{Different loss functions:} 
smooth $ l_1 $, $ l_1 $ and 
$l_{2}$ loss functions show no significant difference,
and $l_{2}$ works slightly better.
}
\label{table:loss}
\end{table}

\begin{table}[t]
\small
\centering
\begin{tabular}{c|c c c|c c c} 
 larger? & AP & AP$_{50}$ & AP$_{75}$ & AP$_{S}$ & AP$_{M}$ & AP$_{L}$ \\
 \hline
  & 30.3 & 53.0 & 31.1 & 14.2 & 33.2 & 43.4 \\
 LK & 31.0 & 52.7 &	31.9 & 14.7 & 33.8 & 44.5 \\
 DC & \textbf{31.8} & \textbf{53.9} & \textbf{32.9} & \textbf{15.9} & \textbf{34.2} & \textbf{45.7}
\end{tabular}
\caption{\textbf{Large receptive field matters:} Improving performance with a larger receptive field.}
\label{table:large}
\end{table}

\begin{table}[t]
\small
\centering
\begin{tabular}{c|c c c|c c c} 
 w/mask & AP$^{bb}$ & AP$^{bb}_{50}$ & AP$^{bb}_{75}$ & AP$^{bb}_{S}$ & AP$^{bb}_{M}$ & AP$^{bb}_{L}$ \\
 \hline
  & 39.6 & 58.2 & 42.7 & 22.5 &	43.4 & 52.1 \\
 \checkmark & \textbf{40.4} & \textbf{58.5}	& \textbf{43.5}	& \textbf{24.5}& \textbf{43.8} & \textbf{52.7}
\end{tabular}
\caption{\textbf{Learning mask boosts object detection:} The performance of detection is advanced by multi-task learning.}
\label{table:results_detection}
\end{table}

\begin{table}[t]
\small
\centering
\begin{tabular}{c|c|c c c|c} 
 Scale & Method & AP & AP$_{50}$ & AP$_{75}$ & FPS \\
 \hline
 416 & ESE-Seg~\cite{xu2019explicit} & 21.6 & 48.7 & 22.4 & \textbf{38.5} \\
 \hline
 400 & \OurMethod & 23.9 & 42.4 & 24.1 & 28.2 \\
 600 & \OurMethod & 28.4 & 49.3 & 28.8 & 18.5 \\
 800 & \OurMethod & \textbf{30.3} & \textbf{53.0} & \textbf{31.1} & 12.8
\end{tabular}
\caption{\textbf{Mask-Based \vs\ Contour-Based:} \OurMethod outperforms ESE-Seg~\cite{xu2019explicit} by a large margin. All models are based on ResNet-50 and the FPS is reported on GTX 1080Ti.
}
\label{table:results_based}
\end{table}

\textbf{Learning Masks boosts Object Detection}
As mentioned in~\cite{fu2019retinamask}, learning with instance mask prediction can usually boost the performance of one-stage detectors. We also find the similar phenomenon in our experiments, \ie, our \OurMethod outperforms FCOS~\cite{tian2019fcos} by $0.8\%$ AP in box, as 
demonstrated 
in Table~\ref{table:results_detection}. Compared with RetinaMask~\cite{fu2019retinamask} which employs %
a few
tricks, our method is %
simpler yet achieving the same performance.
\begin{table*}[t!]

\def\circ{{ - }}

\small
\centering 
\begin{tabular}{ r | r |c c|c c c|c c c}
 Method & Backbone & epochs & aug.\ &   
 AP & AP$_{50}$ & AP$_{75}$
 & AP$_{S}$ & AP$_{M}$ & AP$_{L}$   \\
 \hline
 \textbf{Two-stage} & & & & & & & & & \\
 MNC~\cite{dai2016instance} & ResNet-101-C4 & 12 & $\circ$ & 24.6 & 44.3 & 24.8 & 4.7 & 25.9 & 43.6 
    \\
 FCIS~\cite{li2017fully} & ResNet-101-C5-dilated & 12 & $\circ$ & 29.2 & 49.5 & $-$ & 7.1 & 31.3 & 50.0 
    \\
 Mask R-CNN~\cite{he2017mask} & ResNeXt-101-FPN & 12 & $\circ$ & {37.1} & 60.0 & {39.4} & 16.9 & 39.9 & \textbf{53.5} \\
 \hline
 \textbf{One-stage} & & & & & & & & & \\
 ExtremeNet~\cite{zhou2019bottom} & Hourglass-104 & 100 & $\checkmark$ & 18.9 & 44.5 & 13.7 & 10.4 & 20.4 & 28.3 \\
 TensorMask~\cite{chen2019tensormask} & ResNet-101-FPN & 72 & $\checkmark$ & {37.1} & 59.3 & {39.4} & 17.4 & 39.1 & 51.6 \\
 YOLACT~\cite{bolya2019yolact} & ResNet-101-FPN & 48 & $\checkmark$ & 31.2 & 50.6 & 32.8 & 12.1 & 33.3 & 47.1 \\
 PolarMask~\cite{xie2019polarmask} & ResNet-101-FPN & 12 & $\circ$ & 30.4 & 51.9 & 31.0 & 13.4 & 32.4 & 42.8 \\
 PolarMask~\cite{xie2019polarmask} & ResNeXt-101-FPN & 12 & $\circ$ & 32.9 & 55.4 & 33.8 & 15.5 & 35.1 & 46.3 \\
 \hline
 \textbf{\OurMethod} & ResNet-101-FPN &  12 & $\circ$ & 33.0 & 56.4 & 34.0 & 15.2 & 35.3 & 46.3 \\
 \textbf{\OurMethod} & ResNeXt-101-FPN & 12 & $\circ$ & 35.5 & 59.7 & 36.7 & 17.5 & 38.0 & 49.0 \\
 \textbf{\OurMethod} & ResNet-101-FPN-DCN & 12 & $\circ$ & 34.9 & 58.8 & 36.0 & 16.3 & 37.0 & 49.6 \\
 \textbf{\OurMethod} & ResNeXt-101-FPN-DCN & 12 & $\circ$ & 36.8 & 61.6 & 38.4 & {18.1} & 39.2 & 51.8 \\
 \hline
 \textbf{\OurMethod} & ResNet-101-FPN & 36 & $\checkmark$ & 33.9 & 56.2 & 35.4 & 19.8 &	36.1 & 42.3 \\
 \textbf{\OurMethod} & ResNeXt-101-FPN & 36 & $\checkmark$ & 36.9 & 60.5 & 38.9 & 21.8 &	39.0 & 46.7 \\
 \textbf{\OurMethod} & ResNeXt-101-FPN-DCN & 36 & $\checkmark$ & \textbf{38.2} & \textbf{61.7} & \textbf{40.4} & \textbf{22.6} & \textbf{40.0} & 49.3
\end{tabular}
\caption{\textbf{Instance segmentation} \textit{mask} AP on the COCO \texttt{test-dev}. 
Here ``aug." denotes data augmentation, \eg, multi-scale. $\checkmark$ means training with ``aug."
}
\label{table:SOTA}
\end{table*}

\textbf{Mask-Based \vs\ Contour-Based} We compare \OurMethod\ 
against
the recent contour-based method termed ESE-Seg~\cite{xu2019explicit}. To make this a fair comparison, we do not apply any deformable convolutions in our model. As shown in Table~\ref{table:results_based}, \OurMethod shows a large gain compared to the ESE-Seg method. Additionally, when the input scale becomes smaller (\eg, 400), our model still achieves a better performance at a real-time speed. Note that we do not 
specifically 
train a new model here. It indicates that \OurMethod can not only %
achieve good performance in mask AP,
but also 
shows promises 
for real-time applications.  
Besides the performance, our mask-based method also 
shows 
a detail-preserving advantage that ESE-Seg lacks, which %
is
illustrated in Figure~\ref{fig:hollow}.
Experiments demonstrate that the proposed method 
enjoys 
desirable 
properties 
comparing with contour-based algorithms such as PolarMask~\cite{xie2019polarmask} and ESE-Seg~\cite{xu2019explicit}.

\subsection{Comparison with State-of-the-art Methods}
We evaluate \OurMethod on COCO \texttt{test-dev} and compare our results with 
some 
state-of-the-art methods, including both one-stage and two-stage models. The results are shown in Table~\ref{table:SOTA} and Figure~\ref{fig:Visualization}. 
Without 
bells and whistles, \OurMethod achieves a mask AP of $36.9\%$,
which outperforms most one-stage methods by a large margin. Note that we do not use any tricks in our experiments, \eg, auxiliary semantic segmentation supervision. 
Our performance 
may be further 
improved 
with those tricks. 
Moreover, the gap between TensorMask~\cite{chen2019tensormask} and ours is mainly because 1) Tensormask uses a very long training schedule, as well as 2) bipyramid and aligned representation. Considering that these modules are time- and memory-consuming, we do not plug them into our model.

\subsection{Advantages and Limitations}
\OurMethod has the capacity to 
better deal with
 ``disjointed" objects.  
 An example can be found in Figure~\ref{fig:Visualization} (row 3 column 1).

An interesting phenomenon is that, \OurMethod\  surpasses Mask R-CNN~\cite{he2017mask}
when the detected object is small ($21.8\%$ \vs $16.9\%$) while performs worse when the object becomes larger ($46.7\%$ \vs $53.5\%$). We argue that the main reasons are two folds:
\begin{itemize}

    \itemsep -0.1cm

\item  For small objects, the capacity of the single feature vector in our work is not a problem. While in Mask R-CNN, it requires 
the mask prediction head to label 
each pixel of a small object, which is challenging when the object is very small. That is why we outperform Mask R-CNN for small objects.

\item  
As for 
large objects, 
a compact representation vector  is difficult to 
accommodate 
all the details of the mask. 
In this case, non-parametric pixel labelling shows advantages. 
Additional modules to encode details are needed in this case. 

\end{itemize}

\section{Conclusion}

In this work, we have introduced a new, simple single-shot instance segmentation framework termed \OurMethod. Different from previous works that typically solve mask prediction as binary classification in a spatial layout, \OurMethod represents the mask with a fixed-dimensional and compact vector, and casts the task into a %
regression task. The reformation allows the challenging task to be solved by appending
a parallel regression branch to existing one-stage object detectors.
Experimental analyses demonstrate that the proposed framework achieves 
competitive 
accuracy and speed among one-stage paradigms. 
In the future, we will explore the possibility of using other dictionary learning methods for encoding instance masks, and the possibility of applying this idea to other 
instance recognition tasks.

{\small
\bibliographystyle{ieee_fullname}
\bibliography{egbib}
}

\end{document}